\documentclass[journal]{IEEEtran}
\usepackage{amsmath,amsfonts}
\usepackage{array}
\usepackage[caption=false,font=normalsize,labelfont=sf,textfont=sf]{subfig}
\usepackage[colorlinks,urlcolor=blue,linkcolor=blue,citecolor=blue]{hyperref}
\usepackage{hyperref}
\usepackage{textcomp}
\usepackage{stfloats}
\usepackage{url}
\usepackage{amsmath}
\usepackage{verbatim}
\usepackage{graphicx}
\usepackage{booktabs}
\usepackage{algorithm}
\usepackage{algpseudocode}
\usepackage{stackengine}
\usepackage{multirow}
\def\BibTeX{{\rm B\kern-.05em{\sc i\kern-.025em b}\kern-.08em
    \kern-.1667em\lower.7ex\hbox{E}\kern-.125emX}}
\usepackage{balance}
\stackMath

\begin{document}
\title{De-Fake: Style based Anomaly Deepfake Detection}
\author{Sudev Kumar Padhi, Harshit Kumar, Umesh Kashyap and Sk. Subidh Ali
\thanks {Sudev Kumar Padhi, Harshit Kumar, Umesh Kashyap and Sk. Subidh Ali are with the Indian Institute of
Technology Bhilai, Durg 491002, India (e-mail: sudevp@iitbhilai.ac.in; subidh@iitbhilai.ac.in).}}

\markboth{ }%
{How to Use the IEEEtran \LaTeX \ Templates}

\maketitle

\begin{abstract}
 Detecting deepfakes involving face-swaps presents a significant challenge, particularly in real-world scenarios where anyone can perform face-swapping with freely available tools and apps without any technical knowledge.  Existing deepfake detection methods rely on facial landmarks or inconsistencies in pixel-level features and often struggle with face-swap deepfakes, where the source face is seamlessly blended into the target image or video.  The prevalence of face-swap is evident in everyday life, where it is used to spread false information, damage reputations, manipulate political opinions, create non-consensual intimate deepfakes (NCID), and exploit children by enabling the creation of child sexual abuse material (CSAM). Even prominent public figures are not immune to its impact, with numerous deepfakes of them circulating widely across social media platforms. Another challenge faced by deepfake detection methods is the creation of datasets that encompass a wide range of variations, as training models require substantial amounts of data. This raises privacy concerns, particularly regarding the processing and storage of personal facial data, which could lead to unauthorized access or misuse.  In this paper, we propose SafeVision,  a privacy preserving face-swap deepfake detection method that leverages the unique style features of individuals to detect discrepancies between real and face-swap images. Every facial image of a person exhibits a distinctive style characterized by consistent visual features that remain relatively stable despite environmental variations like lighting, expressions, pose, age progression, background variations, etc. Our key idea is to identify these style discrepancies to detect face-swapped images effectively without accessing the real facial image.  We perform comprehensive evaluations using multiple datasets and face-swapping methods, which showcases the effectiveness of SafeVision in detecting face-swap deepfakes across diverse scenarios. SafeVision offers a reliable and scalable solution for detecting face-swaps in a privacy preserving manner, making it particularly effective in challenging real-world applications.  To the best of our knowledge, SafeVision is the first deepfake detection using style features while providing inherent privacy protection.
\end{abstract}


\section{Introduction}
In the past, media forgery in images and videos was done manually, using techniques like splicing, cropping, or overlaying parts of images, which often left visible signs of tampering, such as mismatched lighting or seams. These forgeries were easy to detect because the tools available at the time couldn’t blend images seamlessly, making it obvious when something was altered. However, with the development of deep learning techniques like autoencoders, Generative Adversarial Networks ($GANs$), and more recently, diffusion models, creating fake media has become significantly easier, more accessible and more convincing. These technologies can produce highly realistic images and videos, making it increasingly challenging to distinguish between authentic and manipulated content.

Autoencoders play a key role in deepfake creation by encoding and decoding facial features, enabling the manipulation of facial expressions and characteristics~\cite{AE1,AE2,AE3}. GANs take this a step further, being especially popular for creating deepfakes, as they can swap one person’s face with another in a video, often producing results so realistic that they can fool even experts~\cite{GAN1,GAN2,GAN3}. More recently, diffusion models have been introduced to enhance the quality of these fakes, particularly in face-swapping, by refining the generated images or videos to make them even more convincing~\cite{diff1,diff2,diff3}. This technique is now widely accessible, as tools and apps have made it easy for anyone to create deepfakes without specialized knowledge. As deepfakes become increasingly common, they are often used for harmful purposes, such as spreading false information, damaging reputations, or manipulating political opinions, all of which have serious consequences for privacy and trust. Unfortunately, deepfake technology is also being used to exploit children, enabling the creation of child sexual abuse material (CSAM). Criminals manipulate images or videos of children using AI tools, such as face-swapping onto explicit content or creating entirely fabricated abuse material. They often source photos from social media, transforming harmless images into explicit deepfakes that closely resemble specific children. These manipulated materials are then used for blackmail, coercing victims into producing real explicit content, or even leading to real-world abuse such as molestation or rape. Once created, these recordings are circulated or sold on the dark web, establishing a cycle of exploitation and profit for criminals.

The psychological and social impact on victims is devastating, as this material can remain online indefinitely, perpetually violating their privacy and dignity. Victims are left feeling powerless, and their families endure immense emotional distress. Law enforcement struggles to track and remove this content, particularly in private networks like the dark web. As these technologies continue to improve, it becomes increasingly important to develop effective methods for detecting deepfakes. This includes identifying subtle clues, such as unusual facial movements, inconsistent lighting, or hidden artifacts that are difficult for the human eye to detect. Addressing this growing issue requires stronger laws and advanced detection systems to support victims.


Recently Many methods have been developed to detect deepfake images. However, these models often fail to adapt well, showing weaker performance when tested against newer deepfake or image manipulation techniques. Deepfake detection involves identifying subtle differences between real and synthetic images, often by leveraging semantic visual cues such as abnormal blending boundaries~\cite{dt28-27} or facial incongruities~\cite{dt20}. Early methods~\cite{dt1,dt10,dt37,dt42Faceforensics++} approached this problem as a binary classification task, using deep convolutional neural networks (CNNs) to distinguish real images from fake ones. While these methods performed well on controlled datasets, they were heavily dependent on the distribution of training data and lacked a deep understanding of forgery principles~\cite{dt50}. As a result, their generalization to unseen scenarios was limited.

\begin{figure}
    \centering
    \includegraphics[width=0.9\linewidth]{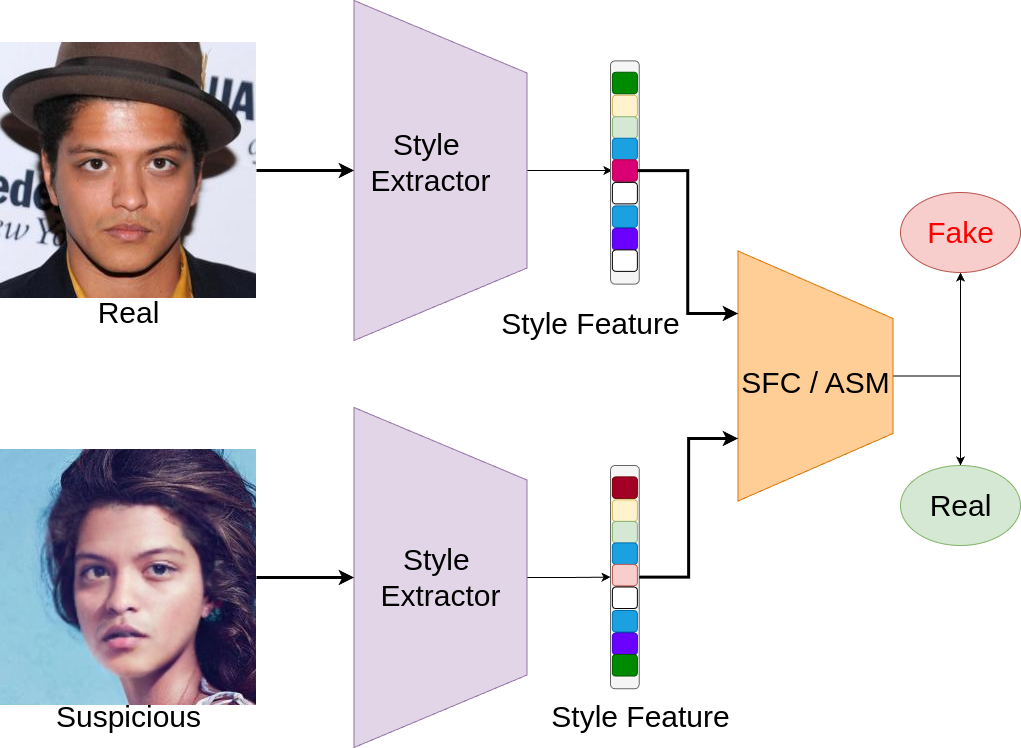}
    \caption{Overview of the Proposed Method}
    \label{fig:over}
\end{figure}

To address these challenges, subsequent research explored more specific forgery patterns, such as noise analysis~\cite{dt28-27}, distinct facial regions~\cite{dt7,dt53}, and frequency-domain features~\cite{dt19,dt41}. Similarly, another line of work focused on domain-specific features like up-sampling artifacts~\cite{dt32} and  key facial areas like the mouth and eyes~\cite{dt13,dt20,dt34}. These advancements have enhanced detection accuracy for certain manipulation techniques. However, as deepfake methods evolve, these visual artifacts have become significantly less perceptible, reducing the reliability of such approaches. Furthermore, methods that focus only on specific regions may overlook valuable clues in other areas, limiting their robustness. To address this limitation, the proposed implicit identity-driven method~\cite{implicit} introduces two types of identity features in a face: explicit identity features, which describe the visual appearance of the face, and implicit identity features, which capture hidden identity details. In real faces, the explicit and implicit identity features are consistent, whereas in fake faces, they are misaligned. Leveraging two specifically designed techniques, the method trains the system to bring real faces closer to their true identity representation while pushing fake faces further away. This approach enhances the system’s ability to detect fake faces, even in challenging scenarios.
However, this technique also suffers from advanced deep face-swapping techniques where target style features are blended properly with a source image.

To address the above issue, we propose a style-based deepfake detection method where we extract style features from different layers of a pre-trained model for both real and corresponding fake images. By embedding these style features into a model, we create a framework that minimizes the distance between a real image of a person and its corresponding fake, while maximizing the distance between real and fake images in general. Each person has a unique style, which remains consistent across various environmental factors like lighting and weather. Even if a source image is blended seamlessly onto a target image to generate a fake, the unique style of the person can still be detected. By comparing the style features of real and fake images, our method can effectively identify whether an image is real or manipulated, making it easier to detect suspicious images in real-world scenarios
\section{ Related Works}
\subsection{Deepfake Creation}

The creation of deepfakes has evolved significantly, driven by advancements in generative models like autoencoders, Generative Adversarial Networks ($GAN$) and diffusion models. Deepfake methods can be broadly categorized into facial reenactment, attribute editing, synthetic face generation, and face swapping, with the latter being a prominent area of focus due to its unique challenges in identity manipulation. A fundamental challenge in face-swapping lies in disentangling identity features, which define who a person is, from transient attributes such as pose, expression, and lighting conditions.

Facial reenactment methods, such as Face2Face~\cite{thies2016face2face}, pioneered real-time facial manipulation by transferring expressions from a source to a target face while preserving the target's identity. These approaches primarily focused on expression transfer, with limited attention to identity manipulation. Similarly, attribute editing frameworks, like StarGAN~\cite{cat}, enabled multi-domain transformation (e.g., age, gender, expression), leveraging disentanglement to some extent but without addressing identity replacement. Synthetic face generation methods, particularly StyleGAN~\cite{style2}, further advanced disentanglement by enabling precise control over style and semantic attributes, laying the groundwork for modern face-swapping techniques.

Face swapping, unlike other categories, uniquely aims to replace the identity in an image or video while maintaining the target's pose, expression, and lighting. Early methods, such as DeepFaceLab~\cite{deepfacelab} and FaceSwap~\cite{faceswap}, utilized autoencoders to reconstruct and blend faces, enabling practical yet limited results. These approaches often struggled with occlusions, extreme poses, and lighting inconsistencies due to inadequate disentanglement of identity and attributes. Tools like Roop~\cite{roop} and Roop Unleashed~\cite{roopu} further simplified the process by offering one-click solutions, emphasizing accessibility. Recent research has introduced explicit disentanglement mechanisms to address these challenges. SimSwap~\cite{simswap} employed feature-matching losses to ensure identity preservation while aligning transient attributes with the target. FaceShifter~\cite{GAN3} extended this idea by introducing adaptive identity embeddings in a two-stage pipeline to enhance realism and fidelity, even in challenging scenarios like extreme poses or occlusions. HiFiFace~\cite{hifi} improved upon these ideas with hierarchical disentanglement, isolating identity features from both low-level textures and high-level semantics, resulting in superior robustness. One-shot frameworks, such as Ghost~\cite{ghost} achieve effective face swapping with minimal training data. Lightweight implementations like MobileFaceSwap~\cite{xu2022MobileFaceSwap} expand the applicability to mobile devices, showcasing efficiency and accessibility. Emerging techniques like StyleSwap~\cite{styleswap} and E4S~\cite{liu2022fine} leverage GAN inversion to empower robust identity preservation and style disentanglement. Diffusion-based approaches, such as DiffFace~\cite{diff2} use facial guidance to achieve high-precision results.

The disentanglement of identity and attributes remains a key area of focus in face-swapping research. While earlier approaches struggled with blending identity and transient features, modern methods explicitly address these challenges using dedicated embedding networks, hierarchical representations, and feature-matching losses. However, challenges persist, particularly in ensuring temporal consistency for video applications and handling extreme environmental variations. These advancements not only improve visual fidelity but also highlight the importance of integrating feature-level disentanglement into face-swapping pipelines for realistic and controllable facial synthesis.

\subsection{Deepfake Detection}

Deepfake detection has evolved rapidly in recent years, encompassing a wide range of techniques, from traditional machine learning and statistical methods to advanced deep learning approaches. With the increasing sophistication of deepfake generation tools, researchers have explored diverse strategies to improve detection robustness and accuracy, often focusing on the challenges of generalization across varied datasets and the need for explainable models. In the domain of image analysis, techniques employed saturation cues to distinguish $GAN$-generated images from real ones~\cite{mccloskey2020saturation}.  Further, a Support Vector Machine ($SVM$) based technique leveraging edge feature detectors like HOG, ORB, and SURF is used to achieve detection accuracy of $95$\% ~\cite{kharbat2019svm}.  XGBoost, a popular tree-based model was also used to exploit visual discrepancies in deepfake generation, such as artifacts or inconsistencies in face manipulation~\cite{ismail2020xgboost}.

Alongside machine learning, statistical techniques have been applied to uncover subtle discrepancies between real and fake data. Photo Response Nonuniformity ($PRNU$), a unique noise pattern generated by camera sensors, has also been explored for detecting deepfakes in videos. \cite{koopman2020prnu} showed that deepfake videos often exhibit subtle inconsistencies in $PRNU$ patterns due to synthetic alterations in the image. Their method, which involves comparing the $PRNU$ of video frames, has proven effective in distinguishing manipulated content from genuine footage. MStatAttack~\cite{hou2022mstatattack} aim to minimize the statistical differences between real and fake images in both the spatial and frequency domains. 

Deep learning has become the dominant paradigm for deepfake detection due to its ability to learn complex feature representations from large datasets. $CNN$ has been employed in various forms to detect facial manipulation and inconsistencies in videos. XceptionNet and CapsuleNet have shown substantial promise in learning robust representations of facial features, effectively detecting manipulated faces \cite{cheng2018xception,sabour2017capsule}. Capsule networks, with their ability to capture spatial relationships, are particularly effective in detecting subtle inconsistencies caused by deepfake algorithms, while XceptionNet, known for its depth and complexity, excels at identifying fine-grained details in facial images. Multi-task learning has also gained traction in recent studies. Techniques like VA-MLP (Variational Autoencoder Multi-Layer Perceptron) and DefakeHop integrate multiple deepfake detection tasks, such as classification, localization, and anomaly detection, into a unified framework. These models have shown improved generalization across different types of deepfake datasets by learning shared representations from multiple tasks \cite{liu2021vamlp, chen2021defakehop}.

In recent years, Vision Transformers (ViTs) have emerged as a strong alternative to traditional CNN-based architectures. CViT (Convolutional Vision Transformer) combines the strengths of both CNNs and Transformers, capturing both local and global features, thus improving the detection of deepfake faces in images and videos \cite{dosovitskiy2021cvit,}. Moreover, research such as Deepfake Catcher explored the efficacy of simple fusion models that combine shallow features from various deep learning models, often outperforming more complex, deep neural networks. These findings challenge the assumption that deeper, more complex models are always better and indicate that simplicity can sometimes yield robust results in deepfake detection~\cite{li2022deepfakecatcher}. There has been a focus on improving the generalization of deepfake detectors to handle the increasing variety of manipulation techniques. Several papers, such as MMNet~\cite{chen2022mmnet} and MINTIME~\cite{wang2023mintime}, have introduced multi-collaboration and multi-supervision frameworks that enable models to detect deepfakes across different manipulation methods, enhancing their robustness and accuracy across diverse datasets.

Invariant Risk Minimization (IRM), explored by Rethinking IRM for Generalizable Deepfake Detection, seeks to reduce the model's dependency on specific datasets, thus improving its ability to generalize to unseen deepfake content~\cite{zhang2022irm}. Similarly, techniques such as latent space augmentation~\cite{zhou2023latent} and style latent flow modeling~\cite{li2023styleflow} have been proposed to make models more resilient to variations in facial attributes, lighting conditions, and video artifacts. Additionally, recent work such as Deepfake Video Detection Using Facial Feature Points and Ch-Transformer~\cite{zhao2023chtransformer} combines spatiotemporal learning with Transformer networks, enabling detection of inconsistencies not just in facial regions but also in dynamic temporal interactions, addressing the challenge of detecting source manipulation in videos.

\subsection{Existing Deepfake Dataset}
There are several existing deepfake datasets that are used for deepfake detection. Most of the techniques focus on generating deepfakes using a limited number of manipulation techniques which are publicly known~\cite{dt42Faceforensics++,yang2019exposing,dang2020detection,korshunov2018deepfakes,narayan2023df} for some cases and unknown for some synthesis techniques~\cite{dfd,dolhansky2020deepfake,li2020celeb}. Existing dataset~\cite{zi2020wilddeepfake} has also used human annotators to label the dataset as real and fake. Recently, a $DF$-$40$~\cite{yan2024df40} dataset has been released that consists of a sample from $40$ different deepfake techniques, including face-swapping, face-reenactment, entire face synthesis, and face editing. Even though these existing datasets provide a wide variety of samples to train the deepfake detection model, there is hardly any dataset that focuses entirely on face-swapping techniques with sufficient diversity and realism. Most datasets include a mix of deepfake types, making it challenging to develop models specifically tailored to detect face-swapping artifacts.  Additionally, the scarcity of datasets emphasizing nuanced and realistic face-swapping scenarios creates a significant gap in preparing models for diverse and evolving real-world challenges.

\section{Proposed Method }

\subsection{Overview}

Generally, face-swap detection is done using a binary classifier by finding subtle inconsistencies or artifacts introduced during the synthesis process. These methods aim to distinguish real images from manipulated ones by analyzing features such as blending boundaries, noise patterns, or abnormal facial expressions. Early approaches heavily relied on labeled datasets, treating the task as a supervised learning problem. While effective on controlled datasets, these models often struggled to generalize to deepfakes generated by unseen techniques or novel algorithms. This limitations are tackled to a certain extent using  noise analysis and frequency-domain features for spotting hidden artifacts. Although these advancements improved accuracy for certain manipulation methods, the continuous refinement of face-swap  methods has made such artifacts harder to detect. 

To address these challenges, we introduce our style feature-based method for face-swap detection. We have employed two distinct techniques for face-swap detection. The first approach builds a classifier specifically trained to distinguish between real and face-swapped images, leveraging a supervised learning framework. The second approach incorporates an anomaly detection module, which identifies inconsistencies or outliers in facial features and attributes, flagging them as potential face swaps on their deviation from the normal distribution of real images. These two complementary methods improve the robustness and accuracy of face-swap detection.

\subsection{Problem Formulation}
In all face-swapping approaches, the goal is to replace the face of the target individual with that of the source individual in a given image or video.  The generated face-swapped image should have the identity of the source individual, while the facial expression, pose, and lighting condition correspond to the image of the target individual. This concept is analogous to neural style transfer~\cite{neusty1,neusty2,neusty3}, where the style of one image is merged with the content of another to create a new image. In the context of face-swapping, the "style" represents the identity of the source person, while the "content" encompasses the facial expression, pose, and lighting conditions. If there are two normal images of the source person, they are likely to exhibit greater similarity in style compared to the real source image and a face-swapped image, as the latter introduces inconsistencies due to the manipulation. This disparity can be utilized to develop a face-swap detection technique that analyzes the differences between two images: the original source image and the suspicious image (allegedly face-swapped). In our proposed method, we use two images as input for our model to identify the face-swapping. The first image is the suspicious image which is doubted to be face-swapped and the second image is a real image of the source individual.

\subsection{Style-Based Face-Swap Detection as a Classification Problem }
This technique consists of two main modules: the Style Feature Extractor ($SFE$) and the Style Feature-based Classifier ($SFC$). The $SFE$ utilizes a pre-trained image recognition model to process a pair of images, one suspected to be a face-swapped image and the other corresponding to the original image. The $SFE$ extracts multi-layer features from both images by capturing outputs from different layers of the pre-trained model. These layer-wise features are concatenated from selected layers to form a comprehensive feature vector.

After concatenation, the features from all selected layers are stacked together to form a single feature vector, which is then used to train the classifier to detect whether the input image is a face-swapped or a real image. By learning the distinctions between the stacked features of real-real and fake-real pairs, the $SFC$ effectively detects face swaps. The following subsections provide detailed explanations of each module.
\begin{figure*}
    \centering
    \includegraphics[width=0.9\linewidth]{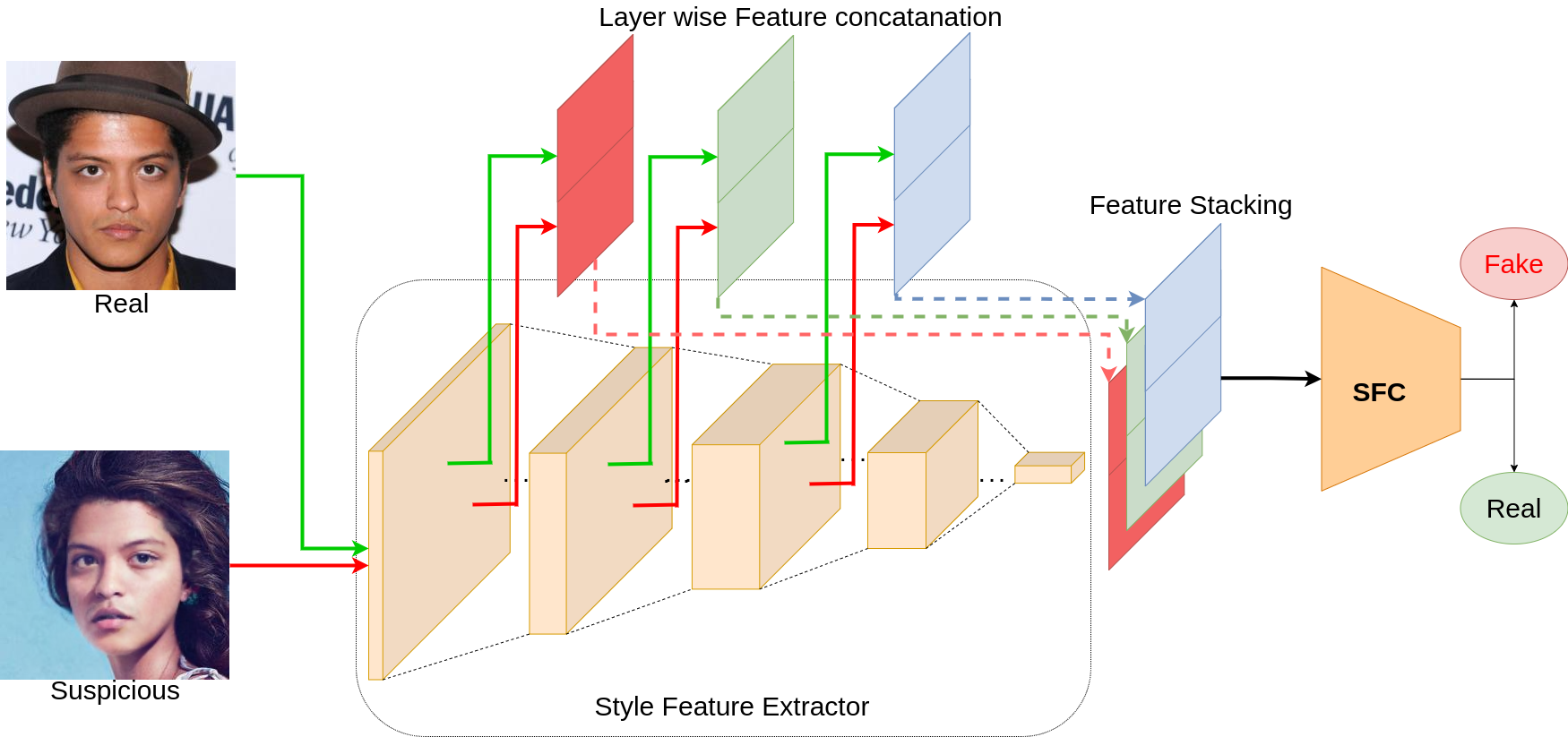}
    \caption{Overview of the Proposed Method}
    \label{fig:over}
\end{figure*}

\subsubsection{Style Feature Extraction}

Since the face-swapped image derives its appearance from both the source and target faces, we hypothesize that the face-swapped image contains distinguishable stylistic features from the original source face. To capture this, we propose to extract and compare the style features from different layers of a pre-trained face recognition model to highlight the differences between real and fake images. Specifically, given an aligned face image $x^i \in \mathbb{R}^{h \times w \times c}$ (where $h$, $w$, and $c$ represent the height, width, and number of channels of the input image), we use a pre-trained image recognition model to extract features from multiple layers, denoted as $SFE_l(x^i)$, where $l$ corresponds to different layers of the network.

We process two input images: one suspected to be a face-swapped image (suspicious) and the other its corresponding original image. The $SFE$ extracts layer-wise features for both images and concatenates them layer by layer to form a comprehensive feature representation for each pair of images. This concatenated representation captures both high-level and low-level differences between the real and fake images. These features are stacked together into a final feature vector, which is passed to the $SFC$.

\subsubsection{Style Feature-based Classifier}

The $SFC$ is designed as a convolution neural network ($CNN$) based classifier to differentiate between real-real and fake-real image pairs. It leverages the multi-layer feature representations extracted by the $SFE$. The $SFE$ extracts and stacks feature vectors from various layers of a pre-trained face recognition model, which are then used as input to the classifier for robust classification. 

To optimize the classification process, the $SFC$ employs a combination of two loss functions: the traditional Binary Cross Entropy ($BCE$) loss and the proposed Stacked Identity Loss ($SIL$). The final loss function used during training is the weighted sum of these two losses, with the $SIL$ scaled by a coefficient $\alpha$.

\subsubsection{Stacked Identity Loss}

The Stacked Identity Loss ($SIL$) is specifically designed to operate across the multi-layer feature representations extracted by the $SFE$. It encourages the classifier to preserve style (identity) consistency for real-real pairs by maximizing the cosine similarity between their feature vectors, while minimizing the cosine similarity for fake-real pairs to capture the stylistic discrepancies caused by face-swapping. The $SIL$ is formally defined as:

\begin{equation} 
    \begin{split} 
    L_{\text{SIL}} = \frac{1}{N} \sum_{i=1}^{N} \sum_{l=1}^{L} \Bigg[ y_i \cdot \left( 1 - \frac{f_{1l}(x_i^{\text{real}}) \cdot f_{2l}(x_i^{\text{real}})}{\|f_{1l}(x_i^{\text{real}})\| \|f_{2l}(x_i^{\text{real}})\|} \right) \\
    + (1 - y_i) \cdot \frac{f_{1l}(x_i^{\text{real}}) \cdot f_{2l}(x_i^{\text{fake}})}{\|f_{1l}(x_i^{\text{real}})\| \|f_{2l}(x_i^{\text{fake}})\|} \Bigg], 
    \end{split}
\end{equation}

where $f_{1l}(x_i^{\text{real}})$ and $f_{2l}(x_i^{\text{real}})$ represent the feature vectors extracted from the $l$-th layer of the pre-trained model for two real images in the real-real pair, while $f_{1l}(x_i^{\text{real}})$ and $f_{2l}(x_i^{\text{fake}})$ represent the feature vectors for the real and fake images in the fake-real pair.

\subsubsection{Binary Cross Entropy Loss}

The $BCE$ loss is used to penalize the deviation between the classifier’s predicted probabilities and the ground truth labels. It helps the classifier minimize classification error and improve accuracy for both real-real and fake-real pairs. The $BCE$ loss is defined as:

\begin{equation} 
    L_{\text{BCE}} = - \frac{1}{N} \sum_{i=1}^{N} \left[ y_i \log(\hat{y}_i) + (1 - y_i) \log(1 - \hat{y}_i) \right],
\end{equation}

where $\hat{y}_i$ is the predicted probability for image pair $i$ (i.e., the classifier’s confidence that the pair is real-real or fake-real), $y_i$ is the ground truth label ($1$ for real-real, $0$ for fake-real), and $N$ is the number of image pairs in the batch.

\subsubsection{Final Classification Formulation}

The final loss function used for training the classifier is the weighted combination of the $BCE$ loss and the $SIL$, scaled by a coefficient $\alpha$. It is defined as:

\begin{equation} 
    L_{\text{final}} = L_{\text{BCE}} + \alpha \cdot L_{\text{SIL}}.
\end{equation}

The classifier detects a wide range of face-swapping manipulations by utilizing both low-level texture features and high-level identity abstractions, leading to improved classification performance. This multi-layer approach ensures the model captures both local distortions and global inconsistencies, resulting in a comprehensive framework for face-swap detection.

\subsection{Style-Based Face-Swap Detection as an Anomaly Detection Problem}

In this approach, face-swap detection is framed as an anomaly detection problem, where the model learns to identify deviations or anomalies in face-swapped images compared to real images. The main goal is to detect unusual or abnormal features that do not align with the expected distribution of real faces, which are indicative of deepfake manipulations. This technique is particularly effective for detecting subtle changes and inconsistencies in the style and identity attributes that are introduced by face-swapping processes. The anomaly detection module comprises two key components: the Anomaly Extraction Network ($AEN$) and the Anomaly Scoring Mechanism ($ASM$). These components work together to capture and quantify these subtle discrepancies between real and face-swapped images.

\begin{figure*}
    \centering
    \includegraphics[width=01\linewidth]{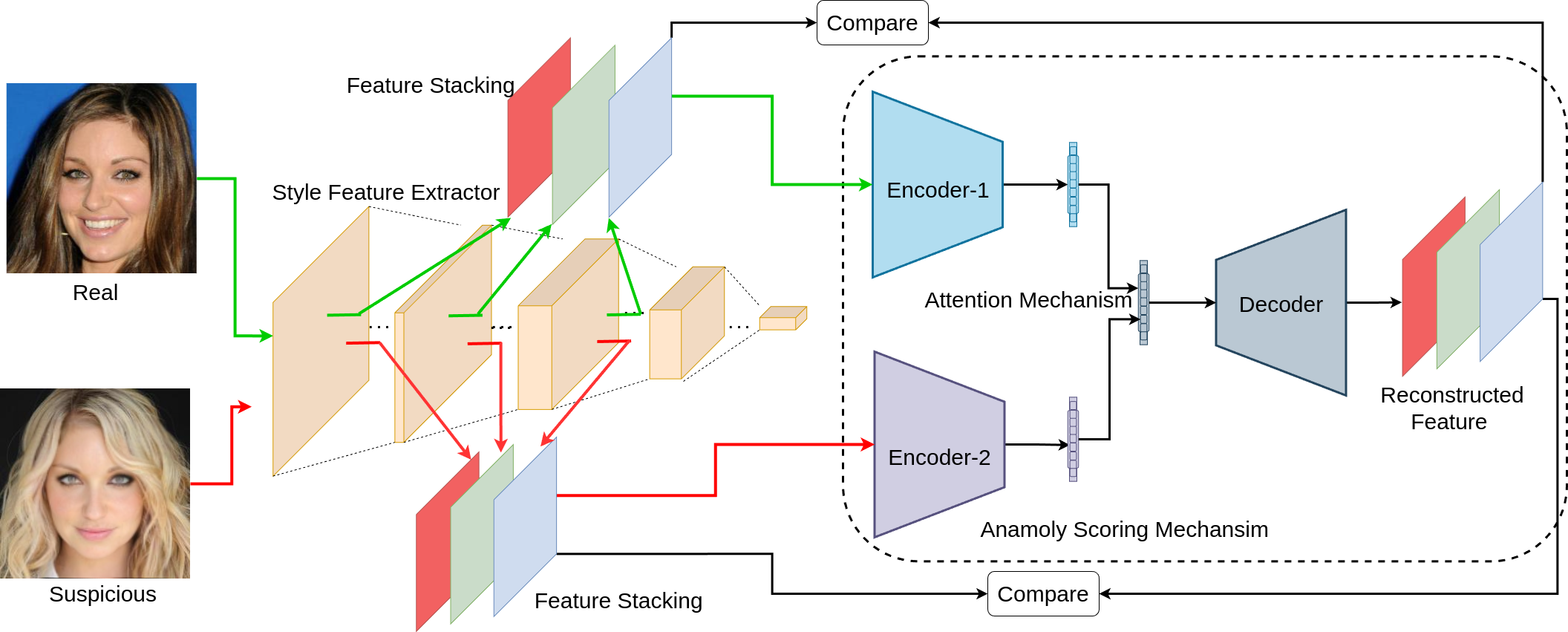}
    \caption{Overview of the Proposed Method}
    \label{fig:anomaly}
\end{figure*}

\subsubsection{Anomaly Extraction Network}

The Anomaly Extraction Network ($AEN$) is responsible for learning the normal distribution of facial images by extracting deep style features. These features capture the unique characteristics of identity and appearance that are present in real images. The $AEN$ utilizes a pre-trained image recognition model, similar to $SFE$ used in the classifier-based approach, to extract features from various layers of the network. These features are then processed through an unsupervised learning framework, enabling the $AEN$ to learn a representation of what constitutes a “normal” image and to detect deviations from this normality.

Given a pair of images, one suspected to be a face-swapped image (suspicious) and the other the corresponding real image, both images are passed through the $AEN$ to extract multi-layer style features. For an image $x^i \in \mathbb{R}^{h \times w \times c}$, the $AEN$ extracts style features $f_l(x^i)$ from multiple layers of the pre-trained image recognition model, where $l$ corresponds to the layer number. 

The $AEN$ processes both images, the real image and the suspicious image, to extract their respective feature vectors. Feature vectors of both images are then stacked separately and passed to $ASM$, which compares these features to highlight differences between the real and suspicious images. Since the $ASM$ has learned the distribution of real images during training, it can identify any discrepancies in the suspicious image that are indicative of face-swapping manipulations.

\subsubsection{Anomaly Scoring Mechanism}

The Anomaly Scoring Mechanism ($ASM$) quantifies the degree of abnormality in a suspicious image by comparing its feature representation to the real image's feature representation. This mechanism is designed to compute an anomaly score that measures how far the suspicious image deviates from the expected style distribution of real images. We use a reconstruction-based loss function to train the anomaly detection framework that minimizes the difference between the reconstructed style (identity) features of the real images and the original features. The goal is to ensure that the $ASM$ captures the normal distribution of real images and detects abnormal features introduced by face-swapping. For reconstruction, we use an autoencoder, which learns to encode the style features into a latent space and subsequently reconstructs them to identify any discrepancies in the features.

We extend the approach of a simple autoencoder by combining a dual encoder architecture with an attention mechanism for more effective anomaly detection. The dual encoder processes the style features of the real and suspicious images, generating two latent vectors, which are then fused using an attention mechanism. The fused latent vector is passed to the decoder, which reconstructs a single feature vector for both images. The model compares the reconstruction errors of the real and suspicious images with the reconstrcuted features, and a high reconstruction error for the suspicious image indicates the presence of face-swapping.

where $f_{\text{reconstructed}}(x^i)$ is the feature vector reconstructed by the $ASM$, and $f(x^i)$ is the original feature vector of the image. By minimizing this loss, the model learns to represent real images and detect anomalous features introduced by face-swapping accurately.

The dual encoder network processes the style features of both images $x_1$ and $x_2$ through separate encoders, resulting in latent vectors $\mathbf{z}_1$ and $\mathbf{z}_2$. These vectors are then fused using an element-wise multiplication (Hadamard Product), which is useful for refining common features between two vectors and generating a combined latent vector $\mathbf{z}_{att}$:
\begin{equation} 
    \mathbf{z}_{att} = \mathbf{z}_1 \odot \mathbf{z}_2
\end{equation}

where $\odot$ denotes the Hadamard product (element-wise multiplication), the updated latent vector $\mathbf{z}_{att}$ is then passed to the decoder to reconstruct the single style feature for both the images.

The final reconstruction loss is computed as the sum of the errors between the reconstructed features $\hat{\mathbf{x}}$  and the original style features:
\begin{equation} 
    L_{\text{recon}} = \| \mathbf{x}_1  - \hat{\mathbf{x}} \|_2^2 + \| \mathbf{x}_2 - \hat{\mathbf{x}} \|_2^2.
\end{equation}

By minimizing this loss, the model learns to reconstruct the normal style features of real images accurately. While testing, we set a threshold of the reconstruction loss, and if the reconstruction loss exceeds the threshold, we can confirm that the suspicious image is a face-swapped image.  This enables the model to effectively detect face-swap manipulations, including face-swapping, even when the manipulations are subtle and do not involve large-scale changes to the facial features.

\section{Experimental Setup}
In this section, we describe the experimental setup used to evaluate our proposed face-swap detection methods. Both approaches are trained using our custom dataset, which contains real and fake image pairs. The dataset consists of aligned face images and is carefully curated to ensure that the face-swapping process reflects realistic manipulations.

\subsection{Creation of Face-Swap Dataset}
Face-swapping is more critical than face synthesis, face reenactment, and editing due to its ability to fully impersonate identities, facilitating misuse of identity theft, fraud, and disinformation with far-reaching societal and ethical consequences. To address this issue, we have created a dataset for face-swap detection. Existing datasets primarily focus on providing face-swapped samples for binary classification, restricting the detection models to differentiate only between real and fake images without addressing more nuanced aspects like manipulation artifacts or identity consistency. To overcome this challenge, we have constructed a dataset that includes the real images of source faces alongside their face-swapped counterparts (where the face present in the target image is swapped with the face of the source image), allowing detection models to effectively analyze identity impersonation and enhance their accuracy in detecting manipulations. Using the source image and the face-swapped image for detection makes sense, as in real-time scenarios, we rely on comparing the original and manipulated content to identify inconsistencies and ensure authenticity. For creating the face-swapped dataset, we have used $18$ different face-swapping techniques to generate the fake content as shown in Table~\ref{table:data}.  We have used a combination of the $CelebA$~\cite{celeba} dataset and the $VoxCeleb2$~\cite{vox} dataset to create the face-swapped images. We have covered multiple scenarios that simulate real-world applications and challenges in face-swapping including variations in lighting, angles, expressions, and other factors to reflect realistic face-swap challenges. We have labelled the real image of the source as real and the face-swapped image as suspicious in the dataset. 

\begin{table}[!htb]
\centering
\begin{tabular}{|c|c|c|}
\hline
\textbf{Face-Swapping Technique} & \textbf{Model Type} & \textbf{Dataset Used} \\ \hline
FaceSwap~\cite{faceswap}         & GAN-based           & CelebA~\cite{celeba}       \\ \hline
SimSwap~\cite{simswap}          & GAN-based           & CelebA~\cite{celeba}       \\ \hline
InSwapper~\cite{inswapper}      & Encoder-Decoder     & CelebA~\cite{celeba}       \\ \hline
E4S~\cite{liu2022fine}          & Encoder-Decoder     & CelebA~\cite{celeba}       \\ \hline
REface~\cite{diff2}             & GAN-based           & Custom~\cite{customdataset} \\ \hline
Facedancer~\cite{facedancer}    & GAN-based           & CelebA~\cite{celeba}       \\ \hline
Mobile Faceswap~\cite{xu2022MobileFaceSwap} & GAN-based     & CelebA~\cite{celeba}       \\ \hline
Roop~\cite{roop}                & GAN-based           & CelebA~\cite{celeba}       \\ \hline
BlendFace~\cite{blendface}      & GAN-based           & CelebA~\cite{celeba}       \\ \hline
Ghost~\cite{ghost}              & GAN-based           & Custom~\cite{customdataset} \\ \hline
Roop Unleashed~\cite{roopu}     & GAN-based           & CelebA~\cite{celeba}       \\ \hline
FaceShifter~\cite{GAN3}     & GAN-based           & CelebA~\cite{celeba}       \\ \hline
Uniface~\cite{xu2022designing}  & GAN-based           & CelebA~\cite{celeba}       \\ \hline
HifiFace~\cite{hifi}            & GAN-based           & CelebA~\cite{celeba}       \\ \hline
Diffswap~\cite{diff1}           & GAN-based           & CelebA~\cite{celeba}       \\ \hline
FSGAN~\cite{nirkin2019fsgan}    & GAN-based           & CelebA~\cite{celeba}       \\ \hline
StyleSwap~\cite{styleswap}      & GAN-based           & CelebA~\cite{celeba}       \\ \hline
Diffusion Faceswap~\cite{diff3} & Diffusion-based     & CelebA~\cite{celeba}       \\ \hline
\end{tabular}
\caption{List of different face-swapping techniques used to generate fake datasets, along with their model type and dataset.}
\label{table:data}
\end{table}


\subsection{Creation of Real Dataset}
We have created samples of the real dataset by using the $VoxCeleb2$ dataset. The fake dataset consists of $2$ images for each data point; one is the real image of the source, and the other is the face-swapped image.  Similarly, we have created $2$ images for each data point in the real dataset, where both the images belong to the same person. One is labelled as real and the other is labelled as suspicious (the one which is doubted to be face-swapped). This is done such that while training, the model learns to differentiate between real faces and potential face-swapped ones, enhancing its ability to detect subtle manipulations. 

We ensure that each subset is balanced with an equal distribution across various scenarios, including:
\begin{itemize}
    \item Different facial expressions (smiling, frowning, neutral, etc.)
    \item Variations in lighting conditions (dim lighting, direct light, shadows, etc.)
    \item Different background settings (indoor, outdoor, complex backgrounds, etc.)
    \item Diverse head poses (frontal, left-right rotated, upward and downward tilts)
    \item Synthetic artifacts introduced by different face-swap generation methods (e.g., blurring, unnatural lighting, inconsistent shadows)
\end{itemize}

\subsection{Dataset Composition}

Out of the $18$ techniques, we have used $10$ techniques of $Faceswap$, $SimSwap$, $InSwapper$, $Mobile$ $Faceswap$, $Roop$, $Ghost$, $DiffSwap$, $StyleSwap$, $Blendface$ and $HifiFace$ for testing and rest of the techniques are used for testing our trained model. We have generated $50,000$ image pairs each for the face-swapped dataset ($5,000$ images from each technique) and the real dataset for training.  We organize the dataset into three subsets: 
\begin{itemize}
    \item \textbf{Training Set:} A large portion of the dataset is dedicated to training. This set contains approximately $80$\% of the real and fake data, which are balanced in terms of the number of samples and categories (real vs. fake).
    \item \textbf{Validation Set:} The validation set contains $20$\% of the dataset and is used to evaluate the model’s performance during training. It is similarly balanced between real and fake samples.
    \item \textbf{Test Set:} The test set is composed of the data created from the remaining $8$ techniques. A total of $16,000$ image pairs are generated and used for the final evaluation. This set also maintains a balanced distribution of real and fake data and includes challenging examples of face-swap generated from different methods.
\end{itemize}

\subsection{Preprocessing}
Before training and evaluation, all images are preprocessed to ensure consistency and high quality. The preprocessing steps include:
\begin{itemize}
    \item \textbf{Face Detection and Alignment:} We use the $MTCNN$~\cite{mtcnn} face detector to detect faces in the images and videos. Detected faces are then aligned using facial landmarks to ensure consistent orientation for model training.
    \item \textbf{Resizing and Normalization:} All images are resized to a uniform resolution of $256 \times 256$ pixels. Pixel values are normalized to the range [-1, 1] to facilitate the training process.
    \item \textbf{Augmentation:} To increase the robustness of the model, we apply various data augmentation techniques, including random horizontal flipping, rotation, color jittering, and slight scaling. These augmentations are applied randomly during both the training and validation phases.
\end{itemize}

\subsection{Style-Based Face-Swap Detection as a Classification Problem}

In the classification approach, the objective is to build a model that can classify image pairs as either real-real or fake-real, where the fake-real pair consists of a real image and a face-swapped image. The architecture for this method consists of the following modules:

\subsubsection{Style Feature Extraction}

For effective feature extraction, we employ a combination of robust pre-trained models, each selected to capture distinct aspects of the image. These models include VGG-19~\cite{simonyan2014very}, ResNet-101~\cite{he2016deep}, U-Net~\cite{ronneberger2015unet}, LinkNet~\cite{chaurasia2017linknet}, ArcFace~\cite{deng2019arcface}, VGGFace2~\cite{cao2018vggface2}, CosFace~\cite{cosface}, and FaceNet~\cite{facenet}  which are chosen for their ability to extract both high-level semantic and low-level stylistic features. Table~\ref{table:rec} summarises the different recognition models used to extract the style features. This enables a comprehensive comparison between real and face-swapped images. Both the real and suspicious (face-swapped) images are processed through these models, which extract multi-layered features. These extracted features are then concatenated to form a unified representation for subsequent analysis and classification.

\begin{table}[!htb]
\centering
\begin{tabular}{|c|c|c|c|}
\hline
\textbf{Model} & \textbf{Domain} & \textbf{ Accuracy (\%)} & \textbf{Technique} \\ \hline
VGG-19 &  Classification & 71.3\%  & Deep CNN  \\ \hline
ResNet101 &  Classification & 77.3\%  & Skip connections \\ \hline
U-Net &  Segmentation & 85\% & Encoder-decoder \\ \hline
LinkNet &  Segmentation & 80.4\% & Encoder-decoder  \\ \hline
ArcFace & Face Recognition & 99.8\% & Angular Margin Loss  \\ \hline
VGGFace2 & Face Recognition & 99.0\%  & Deep CNN  \\ \hline
CosFace & Face Recognition & 99.6\%  & Cosine Margin Loss \\ \hline
FaceNet & Face Recognition & 99.63\%  & Triplet Loss \\ \hline
\end{tabular}
\caption{Different recognition  models in terms of domain,  accuracy, and technique .}
\label{table:rec}
\end{table}

\subsubsection{Style Feature-based Classifier}
Once the style features have been extracted from each model, they are concatenated to form a comprehensive representation of both the real and suspicious images. This process ensures that both low-level and high-level features from different layers and models are captured. The resulting feature vector is then passed to $SFC$ for binary classification. The multi-layer feature extraction process allows us to capture diverse aspects of the images, from fine-grained texture details to high-level identity and structural differences. The $SFC$ is trained using the following configuration:
\begin{itemize}
 \item \textbf{Architecture}: $5$ layer deep $CNN$.
    \item \textbf{Optimizer}: Adam optimizer with a learning rate of $0.0001$.
    \item \textbf{Loss Functions}: A combination of $BCE$ loss and $SIL$.
    \item \textbf{Batch Size}: $64$.
    \item \textbf{Epochs}: $200$ 
\end{itemize}

\subsection{Style-Based Face-Swap Detection as an Anomaly Detection Problem}

In the anomaly detection approach, the goal is to identify suspicious (fake) images by measuring their deviation from the normal distribution learned from real images. Here, the real images are treated as the "normal" class, and the fake images are treated as anomalies. The architecture consists of the following components:

\subsubsection{Anamoly Feature Extraction}

We use the same pre-trained image recognition model for feature extraction as in the classification approach. The suspicious (fake) and real images are processed through the $AEN$, which extracts style features. The extracted features are then compared to identify discrepancies between the real and suspicious images.

\subsubsection{Anomaly Scoring Mechanism}
The extracted features of both the real image and the suspicious image are passed into two different encoders ($E_1$ and $E_2$), respectively. Both the encoders generate the latent vector $z_1$ and $z_2$, respectively, which are then fused using an attention mechanism. The fused latent vector is passed to the decoder ($D$), which reconstructs a single style feature for both images. If the sum of $L_2$ loss between the real-reconstructed and suspicious-reconstructed features crosses a certain predefined threshold, the suspicious image is confirmed to be face-swapped.  The threshold for anomaly scoring is typically selected by analyzing the distribution of $L_2$ loss values on the validation dataset . The Anomaly Scoring Mechanism  is trained using the following configuration:
\begin{itemize}
 \item \textbf{Architecture}: $3$ layer $CNN$ based encoders ($E_1$ and $E_2$) and $3$ layer $CNN$ based decoder ($D$).
    \item \textbf{Optimizer}: Adam optimizer with a learning rate of $0.0001$.
    \item \textbf{Loss Functions}: A combination of reconstruction loss and anomaly score loss.
    \item \textbf{Batch Size}: $32$.
    \item \textbf{Epochs}: $200$ epochs.

\end{itemize}

During testing, the anomaly scores for each image pair are computed, and if the anomaly score exceeds a certain threshold, the image pair is classified as fake.

\subsection{Implementation Details}

Both methods are implemented using the PyTorch deep learning framework. The pre-trained face recognition model is used as a feature extractor, and custom classifiers and anomaly detectors are built on top of this base model.  The models are trained on a machine with RTX $5000$ GPU, with training time for both approaches being approximately $16$ hours for $200$ epochs.

\subsection{Evaluation Metrics}

To evaluate the performance of both models, we use the following metrics:

\begin{itemize}
    \item \textbf{Accuracy}: The proportion of correctly classified image pairs.
    \item \textbf{Precision}: The proportion of true positive predictions among all positive predictions.
    \item \textbf{Recall}: The proportion of true positive predictions among all actual positive samples.
    \item \textbf{F1-Score}: The harmonic mean of Precision and Recall.
    \item \textbf{Area Under the ROC Curve (AUC)}: A measure of the model's ability to distinguish between real and fake image pairs.
\end{itemize}

These metrics provide a comprehensive evaluation of the detection capability of both methods.

\section{Results and Discussion}

In this section, we present the experimental results of the proposed face-swap detection method.  We evaluate the performance of our two primary approaches using our custom dataset, $FaceForensic$++\cite{dt42Faceforensics++}, $Wild$ $Deepfake$~\cite{zi2020wilddeepfake}, $DF$-$TIMIT$~\cite{korshunov2018deepfakes}, $Celeb$-$Df$~\cite{li2020celeb}, $DF$-$Platter$~\cite{narayan2023df} and $DF$-$40$~\cite{yan2024df40}. We have also shown the results of face-swap detection in well-known state deepfake detection techniques of $XceptionNet$~\cite{dt42Faceforensics++}, $SRM$~\cite{srm},$RECCE$~\cite{RECCE}, $SPSL$~\cite{spsl}, $RFM$~\cite{rfm}, $MesoNet$~\cite{afchar2018mesonet}, $EfficientNet$~\cite{efficientnet}, and $Face$ $X$-$Ray$~\cite{facexray}.

\subsection{Feature Extractor Module}
The style feature is extracted using a pre-trained image classifier model in both the case of $SFE$ and $AEN$. As already mentioned, we have used $8$ different models to extract the style features. Table~\ref{table:feature_extractor_performance} shows the performance of both the classification and anomaly detection methods on the extraction of style features using different models. Style features are extracted from intermediate layers because these layers capture textures and patterns that define an image's style~\cite{neusty1,neusty2,neusty3}. Lower layers focus on simple edges and textures, while higher layers capture more complex patterns. The Gram matrix is computed from these layer activations to encode feature correlations, representing style effectively. Multiple layers are used to balance fine-grained details (lower layers) with broader patterns (higher layers). Therefore, we have stacked the intermediate layers for all the models to get overall style features. In Table~\ref{table:feature_extractor_performance}, we have also mentioned the layers we have chosen for different models. As we can see, the feature extracted from the models trained for facial recognition models outperforms the generic image recognition models. Therefore, we will be using $ArcFace$ as our style feature extractor module. 

\begin{table*}[!htb]
\centering
\begin{tabular}{|c|c|c|c|c|c|c|c|}
\hline
\textbf{Model}            & \textbf{Feature Extractor} & \textbf{Layers Feature Extracted From} & \textbf{Accuracy (\%)} & \textbf{Precision} & \textbf{Recall} & \textbf{F1} & \textbf{AUC} \\ \hline
\multirow{8}{*}{Classification} 
                          & VGG-19~\cite{simonyan2014very} & Conv1\_1, Conv2\_1, Conv3\_1, Conv4\_1, Conv5\_1 & 94.2                  & 0.95         & 0.94       & 0.96              & 0.94            \\ \cline{2-8}
                          & ResNet-101~\cite{he2016deep}  & Layer1, Layer2, Layer3, Layer4                 & 93.8                  & 0.94         & 0.93       & 0.95              & 0.93            \\ \cline{2-8}
                          & U-Net~\cite{ronneberger2015unet} & Encoder Conv5, Decoder Upconv3             & 89.7                  & 0.91         & 0.89       & 0.91              & 0.90            \\ \cline{2-8}
                          & LinkNet~\cite{chaurasia2017linknet} & Encoder Stage3, Decoder Stage3            & 90.2                  & 0.92         & 0.90       & 0.91              & 0.91            \\ \cline{2-8}
                          & ArcFace~\cite{deng2019arcface} & Embedding Layer                                & 97.8                  & 0.98         & 0.97       & 0.98              & 0.97            \\ \cline{2-8}
                          & VGGFace2~\cite{cao2018vggface2} & Conv3, Conv4, Conv5                                           & 96.4                  & 0.97         & 0.96       & 0.97              & 0.96            \\ \cline{2-8}
                          & CosFace~\cite{cosface}        & Embedding Layer                                & 97.2                  & 0.97         & 0.96       & 0.97              & 0.96            \\ \cline{2-8}
                          & FaceNet~\cite{facenet}        & Embedding Layer                                & 97.0                  & 0.97         & 0.96       & 0.97              & 0.96            \\ \hline
\multirow{8}{*}{Anomaly Detection} 
                          & VGG-19~\cite{simonyan2014very} & Conv1\_1, Conv2\_1, Conv3\_1, Conv4\_1, Conv5\_1 & 93.2                  & 0.94         & 0.92       & 0.93              & 0.93            \\ \cline{2-8}
                          & ResNet-101~\cite{he2016deep}  & Layer1, Layer2, Layer3, Layer4                 & 92.8                  & 0.93         & 0.92       & 0.93              & 0.92            \\ \cline{2-8}
                          & U-Net~\cite{ronneberger2015unet} & Encoder Conv5, Decoder Upconv3             & 88.4                  & 0.90         & 0.88       & 0.89              & 0.89            \\ \cline{2-8}
                          & LinkNet~\cite{chaurasia2017linknet} & Encoder Stage3, Decoder Stage3            & 89.3                  & 0.91         & 0.89       & 0.90              & 0.90            \\ \cline{2-8}
                          & ArcFace~\cite{deng2019arcface} & Embedding Layer                                & 96.9                  & 0.97         & 0.96       & 0.97              & 0.96            \\ \cline{2-8}
                          & VGGFace2~\cite{cao2018vggface2} & Conv3, Conv4, Conv5                                            & 95.7                  & 0.96         & 0.95       & 0.96              & 0.95            \\ \cline{2-8}
                          & CosFace~\cite{cosface}        & Embedding Layer                                & 96.2                  & 0.96         & 0.95       & 0.96              & 0.96            \\ \cline{2-8}
                          & FaceNet~\cite{facenet}        & Embedding Layer                                & 95.8                  & 0.96         & 0.94       & 0.95              & 0.95            \\ \hline
\end{tabular}
\caption{Performance comparison of Classification and Anomaly Detection models using different feature extractors and layers for style extraction, including accuracy, AUC, F1-score, precision, and recall.}
\label{table:feature_extractor_performance}
\end{table*}

\subsection{In-Dataset and Cross-Dataset Evaluation}
Through quantitative experiments, we verify the performance of our classification and, namely, detection methods to detect faces. Specifically, we evaluate our models using both in-dataset (images generated from $10$ face-swapping techniques our models are trained on) and cross-dataset (images generated from remaining $8$ face-swapping techniques our models are not trained on) with our custom dataset containing face-swapped images. We have taken $10,000$ fake image pairs and $10,000$ real image pairs to test the models. The $L_2$ loss values on a validation dataset have a mean of $0.5$ and a standard deviation of $0.05$; therefore, the threshold is set at the mean plus two standard deviations, resulting in  $0.2$. This value would be used to classify suspicious images as face-swapped if the $L_2$ loss exceeds $0.2$.

\subsubsection{In-Dataset Evaluation}
 We first evaluate the model on our custom face-swapped dataset. As expected, the accuracy of the classification model is $98$\% while the accuracy of the anomaly detection model is $96$\%. This in-dataset evaluation demonstrates that using our method and extracting style features from two images guides the models to learn more discriminative face-swaps with exploitable forgery features.

\subsubsection{Cross-Dataset Evaluation}
To investigate the generalizability of our model, we conduct cross-dataset experiments. We tested our model on unknown face-swap images and still got the accuracy of $97$\% for classification and $93$\% for anomaly detection. When generalizing to a different dataset, our models maintained robust performance, which indicates that both the models can effectively adapt to new, unseen data while preserving high detection performance across varied face-swapping techniques.

 The results obtained demonstrate that both the classification and anomaly detection methods are effective for deepfake detection, with the classification method yielding higher performance across all metrics. The multi-layer feature extraction, combined with the classifier and Stacked Identity Loss ($SIL$), allows the model to capture both local and global discrepancies between real and face-swapped images, leading to high classification accuracy. While not as effective as the classifier-based approach, the anomaly detection method is a promising alternative, especially in settings where labelled data is scarce. 

\subsection{Evaluation of Our Models on Other Datasets}
To assess the performance of our models on established datasets, we conducted experiments on several well-known datasets, including $FaceForensic$++\cite{dt42Faceforensics++}, $Wild$ $Deepfake$~\cite{zi2020wilddeepfake}, $DF$-$TIMIT$~\cite{korshunov2018deepfakes}, $Celeb$-$Df$~\cite{li2020celeb}, $DF$-$Platter$~\cite{narayan2023df}, $DF$-$40$~\cite{yan2024df40}, and our custom dataset. The results confirmed that models trained on our custom dataset, generalized better to unseen datasets, highlighting the model's robustness. This improvement can be attributed to the diverse face-swapping techniques included while training our models. We have taken $2000$ image pairs from each technique. Although every dataset didn't have a corresponding real image associated with it, we still managed to find the real images by just taking images of celebrities for such datasets. As we can see, both our models achieve more than $90$\% accuracy in most of the cases. Even the precision, recall, F1-score and AUC are more than $90$ for most cases. We have achieved less accuracy for $FaceForensics$++, $Wild$ $Deepfakes$, and $Celeb$-$Df$. The primary reason for it was not able to differentiate between the face-swap images and images generated from another forgery such as face editing, synthesis or reenactment. Therefore, the sample we took for testing consisted of a combination of different deepfakes, yielding less accuracy.

\begin{table*}[!htb]
\centering
\begin{tabular}{|c|c|c|c|c|c|c|}
\hline
\textbf{Model}            & \textbf{Dataset} & \textbf{Accuracy (\%)} & \textbf{Precision} & \textbf{Recall} & \textbf{F1} & \textbf{AUC} \\ \hline
\multirow{7}{*}{Classification} 
                          & Custom Dataset   & 98                  & 0.97         & 0.95       & 0.96              & 0.94            \\ \cline{2-7}
                          & FaceForensics++~\cite{dt42Faceforensics++} & 81.4         & 0.80         & 0.80       & 0.80              & 0.80            \\ \cline{2-7}
                          & Wild Deepfake~\cite{zi2020wilddeepfake} & 83.3         & 0.85         & 0.83       & 0.84              & 0.83            \\ \cline{2-7}
                          & DF-TIMIT~\cite{korshunov2018deepfakes} & 92.8         & 0.93         & 0.91       & 0.92              & 0.92            \\ \cline{2-7}
                          & Celeb-DF~\cite{li2020celeb} & 83.5         & 0.84         & 0.85       & 0.84              & 0.83            \\ \cline{2-7}
                          & DF-Platter~\cite{narayan2023df} & 95.6         & 0.94         & 0.92       & 0.93              & 0.94            \\ \cline{2-7}
                          & DF-40~\cite{yan2024df40} & 94.1         & 0.95         & 0.94       & 0.94              & 0.94            \\ \hline
\multirow{7}{*}{Anomaly Detection} 
                          & Custom Dataset   & 96                  & 0.95         & 0.92       & 0.93              & 0.91            \\ \cline{2-7}
                          & FaceForensics++~\cite{dt42Faceforensics++} & 79.6         & 0.80         & 0.79       & 0.79              & 0.78            \\ \cline{2-7}
                          & Wild Deepfake~\cite{zi2020wilddeepfake} & 81.5         & 0.82         & 0.81       & 0.81              & 0.80            \\ \cline{2-7}
                          & DF-TIMIT~\cite{korshunov2018deepfakes} & 92.6         & 0.93         & 0.91       & 0.92              & 0.91            \\ \cline{2-7}
                          & Celeb-DF~\cite{li2020celeb} & 80.9         & 0.81         & 0.80       & 0.80              & 0.79            \\ \cline{2-7}
                          & DF-Platter~\cite{narayan2023df} & 93.3         & 0.94         & 0.93       & 0.94              & 0.92            \\ \cline{2-7}
                          & DF-40~\cite{yan2024df40} & 93.0         & 0.94         & 0.93       & 0.93              & 0.93            \\ \hline
\end{tabular}
\caption{Performance comparison of Classification and Anomaly Detection models on various datasets, including accuracy, AUC, F1-score, precision, and recall.}
\label{table:model_performance}
\end{table*}

\subsection{Evaluation of Existing Models on Our Dataset}

To test the generalizability of our dataset on different well-known deepfake detection techniques, we have created $10,000$ fake images using all the $18$ techniques. Table~\ref{table:deepfake_techniques} provides the accuracy of well-known techniques on our dataset. We have also included our models for comparison and mentioned the dataset on which other detection methods were trained. From   Table~\ref{table:deepfake_techniques} it is clear that most of the existing deepfake detection techniques which were trained on datasets like FaceForensics++, DFDC and Celeb-DF, show relatively low accuracy on our custom dataset. The accuracy of well-known methods such as XceptionNet (56.7\%), SRM (76.4\%), and RECCE (79.5\%) falls significantly short compared to our models. This demonstrates the challenge of generalizing across different deepfake techniques and datasets. In contrast, our models, Classification (97.3\%) and Anomaly (95.1\%), outperform the existing methods, highlighting our dataset's robustness and superior generalizability and model architectures.

\begin{table}[!htb]
\centering
\begin{tabular}{|c|c|c|c|}
\hline
\textbf{Technique} & \textbf{Dataset} & \textbf{Accuracy (\%)} \\ \hline
$XceptionNet$~\cite{dt42Faceforensics++}          & FaceForensics++        & 56.7 \\ \hline
$SRM$~\cite{srm}                            & Celeb-DF               & 76.4 \\ \hline
$RECCE$~\cite{RECCE}                          & FaceForensics++        & 79.5 \\ \hline
$SPSL$~\cite{spsl}                      & FaceForensics++ & 82.8 \\ \hline
$RFM$~\cite{rfm}                         & Celeb-DF               & 81.2 \\ \hline
$MesoNet$~\cite{afchar2018mesonet}                & FaceForensics++        & 63.3 \\ \hline
$EfficientNet$~\cite{efficientnet}       & DFDC & 59.5 \\ \hline
$Face$ $X$-$Ray$~\cite{facexray}                    & FaceForensics++        & 61.3 \\ \hline
Ours (Classification)                    & Custom        & 97.3 \\ \hline
Ours (Anomaly)                     & Custom        & 95.1 \\ \hline
\end{tabular}
\caption{Comparison of deepfake detection techniques, their models, datasets, and accuracy.}
\label{table:deepfake_techniques}
\end{table}


\subsection{Discussion and Limitations}
One of the important parts of our method is extracting the style features (the identity information), which can be relevant and helpful for deepfake detection. Usually, generic image recognition models such as 
VGG and ResNet excel in feature extraction due to their deep architecture captures complex image features through multiple convolutional layers, making them highly effective for classification and recognition tasks. Similarly, other popular networks, such as U-Net and LinkNet, have encoder-decoder structures, which are adept at capturing both detailed and contextual features, which is particularly important for segmentation. Still, these models fail to capture certain nuance style features when it comes to facial images. Even deep networks such as VGGFace2, which is trained on facial images, fail to capture all the relevant style features.  To address this, we leverage specialized networks designed for facial feature extraction, such as FaceNet, ArcFace, and CosFace, to generate robust facial embeddings via their embedding layers to represent style feature representations that differentiate faces effectively. These embeddings, typically with low-dimensional feature vectors, allow for a compact yet powerful representation of facial characteristics, which are crucial for distinguishing between real and fake faces. This low-dimensional nature of the feature vector ensures efficient computation while maintaining high accuracy in identifying subtle manipulations in facial images, thus enhancing the performance of our face-swap detection methods.

One of the potential drawbacks of our method is that it is prone to failure if the real image provided for style feature extraction is also a deepfake image. We have tested this scenario and observed that our model starts to give random predictions for such inputs, causing the accuracy to drop to $42$\%. As of now, we have not come up with a proper solution to address this issue. However, as an initial approach, we can implement a preprocessing step that identifies potential deepfake images before passing them for style feature extraction. This could involve using a secondary deepfake detection model to filter out deepfake images from the real ones, helping to mitigate the impact on overall performance. Further research is needed to refine this approach and improve robustness in such cases.

\subsection{Conclusion}

In this work, we proposed and evaluated two deepfake detection techniques based on style feature extraction: one using a classifier and the other using anomaly detection. Our experimental results show that both approaches are effective, with the classifier-based method outperforming the anomaly detection method in terms of accuracy and other performance metrics. These findings highlight the potential of style-based features for face-swap detection and open the door for further research into more robust and scalable detection techniques. 
\bibliographystyle{splncs04}
\bibliography{main}

 \end{document}